\documentclass[runningheads]{llncs}
\usepackage[T1]{fontenc}
\usepackage{graphicx}

\usepackage{tabularx}
\usepackage{comment}
\usepackage{enumerate}
\usepackage[shortlabels]{enumitem}
\usepackage{wrapfig}

\usepackage{briefcite}

\usepackage{ifthen}
\newboolean{briefcitations}
\setboolean{briefcitations}{false}  %

\ifthenelse{\boolean{briefcitations}}{
  \includecomment{brief}
  \excludecomment{full}
}{
  \excludecomment{brief}
  \includecomment{full}
}

\begin{document}
\title{Requirements for Recognition \\ and Rapid Response to Unfamiliar Events \\ Outside of Agent  Design Scope}
\titlerunning{Responding Effectively to OODS Situations}

\ifthenelse{\boolean{briefcitations}}{
\authorrunning{Anonymous Review}
}
{
\author{Robert E. Wray\orcidID{0000-0002-5311-8593} \and
Steven J. Jones\orcidID{0000-0003-1942-6354} \and
John E. Laird\orcidID{0000-0001-7446-3241}}%

\institute{Center for Integrated Cognition, IQMRI, Ann Arbor, MI 28105 USA
\email{\{robert.wray,steven.jones,john.laird\}@cic.iqmri.org}\\
\url{http://integratedcognition.ai}}

}

\maketitle              %
\begin{abstract}
Regardless of past learning, an agent in an open world will face unfamiliar events outside of prior experience,  existing models, or policies. Further, the agent will sometimes lack relevant knowledge and/or sufficient time to assess the situation  and evaluate response options. How can an agent respond reasonably to situations that are outside of its original design scope? How can it recognize such situations sufficiently quickly and reliably to determine reasonable, adaptive courses of action? We identify key characteristics needed for solutions, review the state-of-the-art, and outline a proposed, novel approach that combines domain-general meta-knowledge (inspired by human cognition) and metareasoning.  This approach offers potential for fast, adaptive responses to unfamiliar situations, more fully meeting the performance characteristics required for open-world, general agents.
\end{abstract}

\section{Introduction}

General intelligent agents must operate in environments that cannot be fully known.
They will encounter situations outside their past experience, beyond the scope of design or training. Unlike quantitatively unique (\textit{out of distribution}) situations  that align with agent models, these \textit{out-of-design-scope} (OODS) situations are \textit{qualitatively} different. An agent's response to OODS situations cannot be fully prescribed in advance. What should an agent do when it encounters such situations? Abstractly, various properties and desiderata are apparent: survive; minimally impact other people, agents, and the environment; and, when feasible and apt, act so that the tasks and goals prior to encountering the OODS situation can be resumed. Crucially, OODS situations demand and stress agent autonomy: an agent must recognize and respond without specific prior preparation for the specific situation. Further, the agent must operationalize the general desiderata above in the specific but unfamiliar situation in which it finds itself.

Responding to OODS situations is partially addressed by different research approaches, including policy responses for out-of-distribution events in RL \cite{haider_can_2024}, responding to and learning from novelty \cite{mohan_domain-independent_2024,goel_novelgym_2024_local}, and open-world learning \cite{kejriwal_challenges_2024}. This paper explores this challenge, drawing from this prior research as well as perspectives in autonomous agents, AI safety, artificial general intelligence (AGI), and even human cognition. The paper makes three contributions:
\begin{enumerate}
    \item  We identify a set of \textbf{necessary functional requirements of solutions}, more completely instantiating the informal desiderata  above. OODS situations require holistic approaches that take into account the limitations of agent embodiments and agent perseverance in (all) its tasks and goals/mission.
    \item Many methods have been explored; however,  \textbf{analysis herein shows the current states of art fall short of the requirements}. Agents need to recognize and characterize the OODS situation in their specific contexts. 
    {Any pre-defined, fixed response is unlikely to satisfy all requirements. Similarly, any approach that emphasizes fixed dimensions of a problem (such as novelty) is also likely to fall short of the requirements. }
    \item Human behavior  generally meets the requirements. We propose that \textbf{human-inspired \textit{appraisal processes} and domain-general \textit{meta-reasoning strategies}} can enable recognition and (reasonably) adaptive responses to OODS situations in open-world settings.  While human appraisal is often described, theorized, and modeled as a precursor of   
emotion  \cite{roseman_cognitive_1984,scherer_studying_1993,ellsworth_appraisal_2003,frijda_laws_2006,gratch_domain-independent_2004,scherer_emotion_2019},
we emphasize the functional role of appraisal processes in human cognition: appraisal processes produce domain-general meta-knowledge about specific situations that the agent encounters. 
\end{enumerate}

 We hypothesize that appraisal, combined with meta-reasoning, potentially satisfies all  requirements. We do not present an implementation. However, we do identify specific open analytical and empirical questions for future research that would provide human-appraisal-like processing in general intelligent agents.

\section{Responding to Out-of-Design-Scope Situations}

We now define the problem introduced above and contrast with other characterizations of ``unfamiliar'' situations, especially ``out of distribution'' situations researched in reinforcement learning (RL) \cite{haider_can_2024,chen_you_2022,chen_adapt_2023,anand_prediction_2023,hansen_self-supervised_2021}. While the problem characterization is akin to those in the RL literature,  any autonomous agent, regardless of the method used to design and implement it, will likely encounter some situation outside of the scope of its original design and prior experience.

\begin{figure}[tb]
    \centering
    \includegraphics[width=0.75\linewidth]{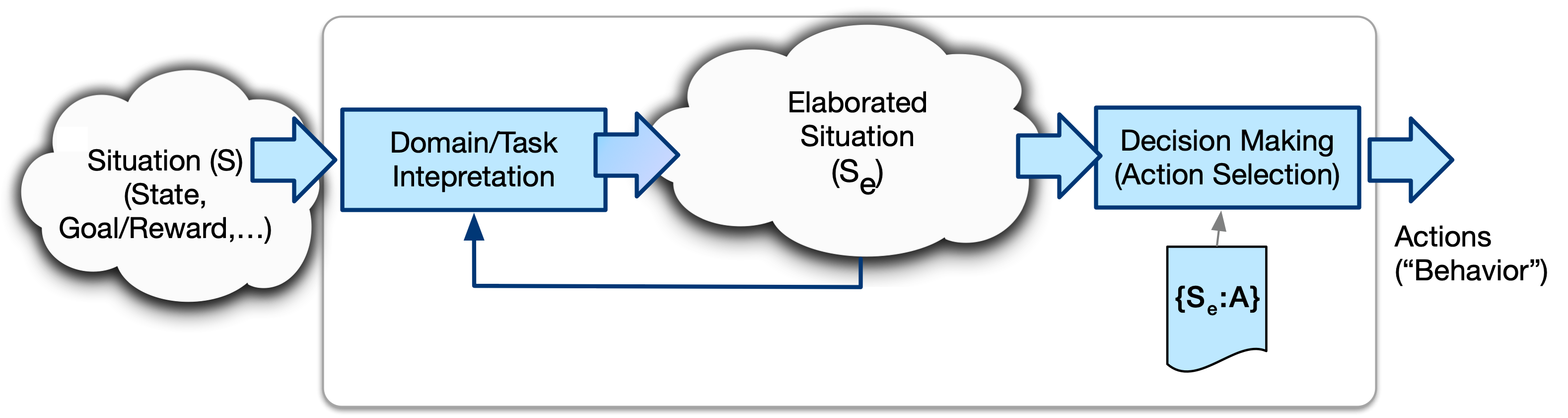}
    \caption{Conceptual illustration of a general/abstract agent decision-making process.}
    \label{fig:strawman-agent}
\end{figure}

To motivate the problem, consider the conceptual agent illustrated in Figure~\ref{fig:strawman-agent}. Agent perceptual processes produce \textit{observations} and, in response to those observations, the agent chooses actions according to a policy that defines an appropriate action.
We do not assume a particular decision-making framework: the decision process could range from choosing actions that optimize reward (similar to RL) to deriving actions based on a world model (as used in classical planning).

While many agents often have policies that directly map observations from the state \textit{S} to actions \textit{A}, more sophisticated agents typically construct or use internal features in decision-making. Examples of internal features include  intermediate representations such as  
hierarchical decomposition for RL \cite{dietterich_hierarchical_2000} or hierarchical task networks \cite{erol_umcp_1994,nau_applications_2005}.
Agent decisions now encompass both the execution of actions and the assertion or updating of internal, elaborated features. In the diagram, we abstract from specific processes by representing the creation of internal features as a unique \textit{elaboration} process, distinct from decisions to take external action. New internal features are derived from current observations and earlier elaborated features. The selection of external actions is thus informed by this elaborated representation of the situation (\textit{$S_e$}), which incorporates both the initial observation/state and any pertinent subsequent elaborations.

Agents can acquire policies that support both elaboration of features and action selection in many ways, including
traditional reinforcement learning \cite{sutton_reinforcement_2018,dietterich_hierarchical_2000}, planning, learning from instruction \cite{laird_interactive_2017}, to even direct knowledge engineering.
The possible situations that an agent might encounter in a realistic, open-world deployment are infinite, but training/development resources (funding, training time, etc.) are finite. Consequently, assumptions about the application domain must be introduced. These design assumptions constrain the scope of possible observations via sensors,  agent affordances, and what features can be elaborated. Consider an autonomous driving system intended for use in North America: a reasonable design assumption could be to train only for right-side driving. These (explicit or implicit) assumptions collectively comprise a \textit{design scope} for any artificial agent. The goal of development and training is to produce an agent that makes appropriate decisions throughout the design scope.

\begin{table}[tbh]
    \centering
    \caption{Categories of Out-of-Design-Scope Situations.}
    \begin{tabular}{p{.1675\columnwidth}p{.82\columnwidth}}
    \hline
    Category & Description \\ \hline \hline
         Out of \newline distribution &  Situation is familiar (observations and elaborated features are apt), but no action is recommended because the specific combination of features (and their values) lies outside the statistical pattern of observations and features experienced previously. \\ 
         \multicolumn{1}{r}{Example:} & Autonomous driver trained on North American roads where the speed limits are typically less than 100 mph. Agent tasked to drive at speeds significantly above 100 mph. The agent was not trained for such speeds and thus lacks  directly (or confidently) applicable policy knowledge.\\ \hline
         Out of \newline designed-feature scope & Observations contain data salient to effective response, but known features are insufficient (or inapt). Agent has some (limited) capacity to recognize and respond, but the design omitted these features/values. \\ 
         \multicolumn{1}{r}{Example:} & Camera-based, lane-following system that was not trained on or exposed to snow being used on snowy roads.  \\ \hline
         Out of \newline observation & Some situations may not be directly perceptible given the agent's embodiment. Without the ability to observe key factors that should influence action, the agent's ability to survive and adapt is quite limited.\\
         \multicolumn{1}{r}{Example:} &  Agent deployed on roads near active volcanoes but lacks perceptual capability to distinguish lava from rocks. Consideration of actions for maneuvering around lava is identical to maneuvering around rocks.  \\\hline
    \end{tabular}
    
    \label{tab:oods_situation_types}
\end{table}

What happens when the agent encounters a situation outside of its policy knowledge (i.e., design scope)? An agent's ability to recognize and respond (appropriately) depends on how the situation interacts with its design scope. Table~\ref{tab:oods_situation_types} summarizes three distinct categories of such situations. Much reinforcement-learning research has been devoted to responding adaptively to 
out-of-distribution situations \cite{haider_can_2024,chen_you_2022,chen_adapt_2023,anand_prediction_2023,hansen_self-supervised_2021}. 
However, much less research attention has been focused on the other two OODS categories in the table.  Nonetheless, deployed systems will encounter situations outside of their design scope. Importantly, the more general, abstract, or open-ended an agent's tasks or missions are, or the more closely it is intended to approach AGI, the more likely OODS situations will be encountered. Imposing \textit{any} design scope for an AGI will likely lead to OODS situations for that AGI system. When these situations arise, without some way to recognize and attempt to respond and to learn from these situations, agents 
will often fail, sometimes with tragic consequences \cite{ntsb_collision_2019}.
Thus, a core problem is the lack of computational approaches that will allow an AGI system to recognize and, when capable, to respond reasonably to out-of-design-scope situations. 

\begin{figure}[bt]
    \centering
    \includegraphics[width=0.75\linewidth]{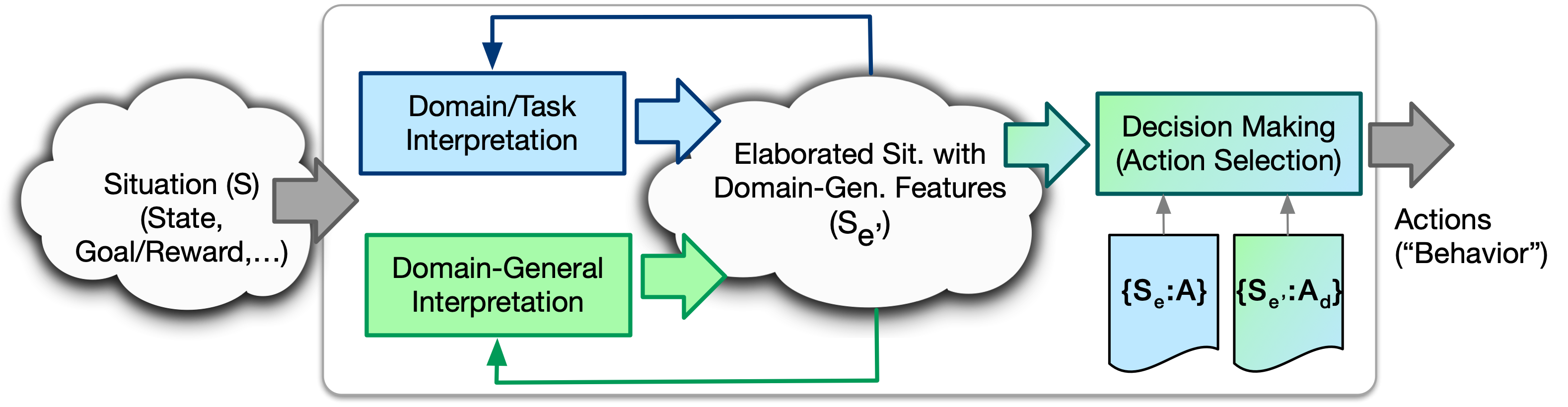}
    \caption{Enhancing agents with domain-general elaboration for OODS situations.}
    \label{fig:compuational-analysis-picture}
\end{figure}

Notionally, we can envision some approach, suggested in Figure \ref{fig:compuational-analysis-picture}, that augments an agent's elaborated situation with domain-general processes that attempt to recognize and characterize OODS situations. 
We assume for now that the resulting elaborated situation $S_{e'}$ is a further, monotonic elaboration of $S_e$ ($\S$\ref{sec:appraisal} outlines possible candidates for a broad and meaningful set of such domain-general features). These supplemental elaborations can then enable (general or default) action responses to OODS situations. In essence, as suggested in the figure, the agent's original policy mapping $S_e\Rightarrow A$ is complemented by a second policy in which the state elaborated with domain-general features, $S_{e'}$, maps to a collection of domain-general decisions/actions $A_d$. Without such capabilities, general agents will be brittle and can fully fail whenever they enter situations outside of their design assumptions.

What requirements must this enhanced decision-making process satisfy to enable agents to thrive in open worlds, where out-of-design-scope situations will repeatedly arise? Table~\ref{tab:requirements} enumerates requirements derived from analyses of others (e.g., \cite{fraifer_autonomous_2025}) and our own experience researching general intelligent agents. These requirements guide subsequent analysis and evaluation of the state of the art and motivate our proposed new approach.

\begin{table}[tbh!]
    \centering
    \caption{Requirements for Responding to   Out-of-Design-Scope Situations.}
    \begin{tabularx}{\linewidth}{cX}
    \hline
     & Description \\ \hline \hline
    R1 & \textbf{Survive (typically) in fail-hard environments}: Situations can present possibly catastrophic outcomes,  due to environment dynamics \cite{kejriwal_challenges_2024}. Thus, it is unacceptable to attempt to adapt to novelty through repeated exposure; a single failure may destroy an agent's embodiment or cause significant damage to the agent or environment. While no agent can ensure it can survive \textit{any} catastrophic event, agents should manifest behavior that generally avoids agent-terminating outcomes.\\ 
    
   R2 & \textbf{Perform in environments with irreversible dynamics}: Resource consumption, elapsed time, and physical changes often make reversing a situation infeasible. Agents must progress while contending with the world they inhabit. General agents, such as physical robots performing active search in an unknown environment compounded by irreversibility, cannot assume the ``safely explorable'' state spaces underlying many AI algorithms  \cite{russell_artificial_1995}. \\ 
  
  R3 & \textbf{Scale to long-duration complex multi-objective tasks}: Adaptation to OODS situations must gracefully scale in three dimensions: time (objectives at different timescales), the complexity of the objective specification (e.g., interpreting laws), and many simultaneous objectives (which may sometimes conflict).  \\ 
    
    R4  & \textbf{Mitigate partial observability}: In realistic situations, an agent's sensors only provide a limited/partial view of the situation. While this requirement appears in many AI systems and algorithms, overcoming or mitigating partial observability is key to OODS problem solving. \\
      
   R5 & \textbf{Adapt to the open world}: General agents cannot assume a closed and stationary environment.  An agent must be capable of both learning new strategies and also adapting/changing existing strategies \cite{kejriwal_challenges_2024}.\\ 
   
   R6 & \textbf{Adapt at any timescale}: Agents should react quickly when the situation demands, but invest more time for further consideration when appropriate. A critical ability for systems that deal with OODS events is their ability to detect such events and successfully mitigate negative effects before the window of opportunity to react to their occurrence has passed  \cite{haider_can_2024}. \\ 
  R7 & \textbf{Persevere with zero-shot adaptation}: Agent must respond to the situation at hand, without assuming the ability to experience a situation multiple times, which is seen as a requirement in continual RL systems  \cite{khetarpal_towards_2022}. Similarly, an agent should persist on its mission during an initial encounter with an OODS situation.\\ 
  
  R8 & \textbf{Improve efficiently}:  OODS events may recur and costs of poor solutions can be significant. Thus, ``muddling through'' subsequent, comparable situations is unacceptable. Agents must improve quickly from a few exposures to such situations. \\\hline

    \end{tabularx}
    \label{tab:requirements}
\end{table}

\section{Analysis of the State of the Art}

We now evaluate several categories of state-of-the-art against the requirements.

\textbf{Prepare with (Broad) Knowledge:} An agent with broad, general knowledge might be able to use that knowledge to respond to an OODS situation. For example, some researchers attempt to generalize pre-existing policy knowledge to make it more applicable in OODS situations \cite{chen_adapt_2023,hansen_self-supervised_2021}. But, given an open world, learned policy knowledge cannot be assumed to be sufficient (complete and correct) for those situations, and thus this approach fails to meet \textbf{R5}. %

\textbf{Extended Deliberation:} %
An agent can use time-intensive (extended) approaches to elaboration and decision making to determine how to respond in OODS situations. Examples include attempting to learn a model appropriate for the situation or replanning (or both \cite{wong_resilient_2018}). Using time-intensive methods violates \textbf{R6}, and, similar to the first item in this section, extended deliberation cannot be assumed to have complete models of the situation sufficient to ensure perseverance (\textbf{R8}) or even survival (\textbf{R1}).

\textbf{Default Fallback:} Another approach is to prepare an agent with general, pre-defined default policies. For example, on encountering an OODS, a UAV may return to base rather than attempt to fulfill its mission (e.g.,  \cite{yang_designing_2020_local}). %
However, a pre-defined fallback fails to improve efficiently (\textbf{R8}), and, if the fallback is to abandon one's tasks, there is no perseverance (\textbf{R7}). Additionally, the success of the fallback depends on design assumptions (what situations to trigger fallback, what the policies should be), so survival and adaptation both depend on design assumptions rather than autonomous capabilities.

\textbf{Return to Familiarity:} In contrast to a fallback, a related approach is to attempt to bias the agent away from ``unknown'' situations toward familiar ones without extended deliberation. One example of this approach in single-life reinforcement learning is to bias an agent towards states with q values to allow it to achieve familiar states and eventually succeed while not assuming direct goal pursuit remains feasible \cite{chen_you_2022_local}. However, attempting to achieve familiar states requires something like reversibility (\textbf{R2}) or at least reachability, and may contravene more direct perseverance of the mission (\textbf{R7}). Additionally, this approach does not itself enable efficient improvement (\textbf{R8}).

\textbf{Learn Slowly:} Some RL agents ignore novelty when first observed, responding with existing policy, as if the situation were familiar. %
For example, a transient value function can be used to capture short-term environmental change to learn to adapt to novel situations \cite{anand_prediction_2023_local}. 
Such gradual adaptation may be acceptable during training but 
does not avoid/mitigate catastrophic failures (\textbf{R1}). %

\textbf{Ensembles and Combinations:} Incomplete solutions to a problem can often be improved by combining prior approaches. Both Mohan et. al. \cite{mohan_domain-independent_2024} and Goel et. al. \cite{goel_neurosymbolic_2024} describe ensemble approaches that first characterize when/how an observation violates or exists outside of an agent's knowledge (i.e., the ``type of novelty'') and then choose how to respond from candidate adaptation methods  (e.g., reprioritizing tasks, replanning, or updating a world model). %
However, to date, this approach chooses actions based only on type of observed novelty. 
While this approach is much closer to a solution than others we have reviewed, it suffers from choosing responses without taking into account situational demands such as time (\textbf{R6}). One situation may demand deliberate modeling and replanning. Another may require a rapid response. Yet another may be informed by a threat to survival (\textbf{R1}). However, the decision process will be the same because the ``type of novelty'' might be the same in those different situations. 

As a second ensemble example, some  applications combine extended deliberation with fallback behaviors. A robot will attempt to replan, but after some predefined search or time limits are exceeded, the robot will revert to a default behavior (e.g., \cite{dumonteil_reactive_2015}). This approach offers significant benefits. Deliberation may lead to apt adaptation at the moment, allowing the agent to continue its mission. When time constraints are exceeded without a solution, an agent may still avoid potentially catastrophic failures with its fallback response. This approach comes close to fulfilling the requirements. However, pre-specified limits are not assured to be responsive to the actual situational demands (\textbf{R6}) and this approach alone does not result in improvement over time (\textbf{R8}), especially when fallback is used.

\section{A New Approach Inspired by Human Cognition}
\label{sec:appraisal}
Thus, while different approaches in the current state of art meet many of the requirements, no existing approach, including ensembles, satisfies the full requirements for responding to and learning from OODS situations. 

However, (adult) human behavior is remarkably resilient in wholly new and unfamiliar situations \cite{bruner_study_2017,book-newell-simon}, including ones with high stakes and time pressure \cite{klein_sources_1998,shortland_choice_2020}. While some of this resilience is attributable to the knowledge and skills adults accumulate over a lifetime \cite{klein_sources_1998}, including the active use of sensemaking and information-gathering strategies when decisions are less time pressured \cite{endsley_toward_1995,klein_making_2006}, various researchers have also identified general, quickly-generated signals, affective states, as being key inputs that modulate how decisions are made \cite{maule_effects_2000,scherer_emotions_2009}. Here we briefly introduce \textit{appraisal theory}, one research paradigm that attempts to explain how such affective states arise in humans. We then outline the general, functional role of appraisals and  hypothesize that similar appraisal-inspired functionality, implemented in general agents, will support improved outcomes in OODS situations,  satisfying all the requirements introduced previously.

\subsection{Theories of (Human) Appraisal Processing}

Theories of appraisal processing arose as a potential explanation for how emotions arise and manifest in human psychology  \cite{roseman_cognitive_1984,scherer_studying_1993}. Aappraisal theories assume a \textit{cognitive} basis for emotion  (e.g., in contrast to theories where emotion is regarded as distinct from cognition, \cite{frijda_emotion_2009}). Thus, a cognitive perspective views appraisal processing as part of the normal process of cognitive decision making, rather than a wholly parallel or meta-process outside of cognition (and volition).

While theories propose different dimensions of appraisal, the overall role of appraisals in these theories is to evaluate the current situation. Examples include familiarity of a situation, its conduciveness to goals and motivations, and assessment of one's control/power to change the situation. In these theories, combinations of appraisal variables lead to different emotional states. For example, someone who experiences something unexpected and out of their control but consistent (conducive) with their current motivation and goals might experience relief \cite{roseman_cognitive_1984}. If the same event was non-conducive, the person might feel fear or discomfort, depending on their certainty of the situation.

A generally less developed aspect of appraisal theories is termed ``action tendencies:'' combinations of specific appraisal variables tend to manifest in certain physical responses \cite{scherer_emotion_2019}. A familiar example is the fight or flight response, but other physical responses are described. For example, when a stimulus is novel and deemed likely goal-relevant, study participants tend to orient or approach that stimulus \cite{scherer_emotion_2019}. 
There have been numerous research efforts focused on computational implementations of appraisal processing, both for psychological modeling \cite{scherer_emotion_2019} and the simulation of realistic emotion in behavior generation \cite{marsella_computationally_2014,gratch_domain-independent_2004,marinier_computational_2009}.

\subsection{The Functional Role of Appraisal Processing}
Other than a precursor for emotion, what is the functional role of appraisal? Individual appraisal dimensions (novelty, pleasantness, urgency, goal-conduciveness, control, normative significance, etc.)  annotate or ``comment'' on how the agent and its cognitive capabilities relate to the external environment. While appraisal generation is informed by the task or domain, these dimensions are not. Collectively, appraisals provide meta-knowledge (or meta-awareness) about general characteristics of situations -- characteristics that human evolution has determined are highly useful for decision-making and adaptation for general agents.

One example of such usefulness is how appraisals inform what decision-making strategies are apt for a situation. A novel, urgent, unpleasant situation  suggests a different decision approach than a novel, non-urgent, pleasant situation. One recurring limitation in existing approaches reviewed above is that the approach was appropriate for some situations but not others: sometimes it is useful to plan; other times, it is better to run away immediately. Appraisals appear to be a rapid, domain-general technology that human metareasoning exploits in selecting an apt decision-making strategy to follow in individual situations.

\begin{figure}[tbh]
    \centering
    \includegraphics[width=0.75\linewidth]{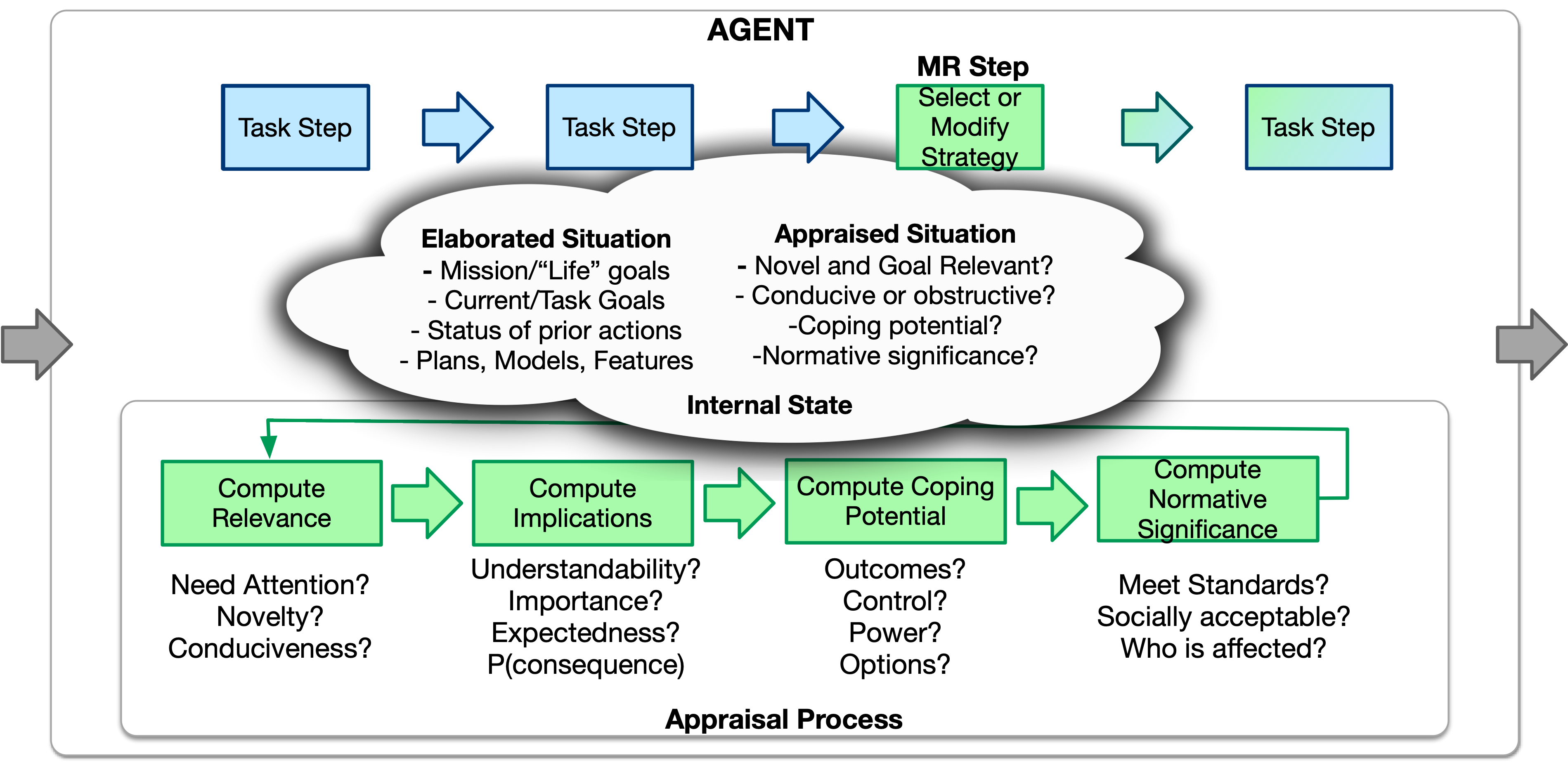}
    \caption{Envisioned approach to using appraisal processing to generate domain-general elaborations (appraisals) to inform agent decision processes.}
    \label{fig:appraisal_approach}
\end{figure}

\subsection{Proposed Approach: Appraisal Processing + Metareasoning}
We hypothesize that a combination of appraisal generation and appraisal-informed metareasoning is sufficient to meet the requirements for responding to OODS situations. Figure~\ref{fig:appraisal_approach} summarizes the envisioned approach. Alongside domain/task reasoning, appraisal processes produce various individual appraisals. The task context informs the generation of individual appraisal instances, but the appraisal processes themselves are not specific to any domain. The set of appraisal variable instances then informs the selection or modification of the task decision strategy (e.g., a ``fast and frugal'' heuristic \cite{gigerenzer_fast_2004_local} vs. a more deliberate one). 

Table~\ref{tab:appraisal_vs_requirements} summarizes an analysis of how this approach can potentially satisfy the requirements. One key advantage is that while appraisal processes are (largely) fixed, appraisal processing is sensitive to the specific OODS situation. Some sets of appraisal variable instances will suggest the agent attend or explore to seek out additional understanding of its situation. In contrast to fallback strategies, appraised characteristics of the situation lead to action choices based on the situation, rather than choosing from pre-defined, fixed policies.

\section{Conclusions}
We explored the problem of artificial, general intelligent agents encountering situations outside of the scope of their design. Every artifact is created under some design assumptions; it seems self-evident that artifacts applied outside of design scope may be less reliable. However, general agents must respond  gracefully and resiliently to these out-of-design-scope situations, ideally surviving, not hurting others or damaging property, and able to resume their tasks. 

After enumerating the requirements for responding to OODS situations, we observed no existing approach met all the requirements. In contrast, humans do seem able to respond resiliently and adaptively in novel situations. Taking inspiration from human behavior, we hypothesized that appraisal processes in agents can provide a set of general characteristics about  situations. Appraisal can inform which decision making strategies are most relevant to a novel, OODS situation, enabling both effective zero-shot coping strategies but also the potential to learn and to improve for these situations in the future.

For the future, we plan to evaluate the overall hypothesis, including empirical assessment of the solution against each of the requirements identified in this paper. Many more fine-grained research and algorithmic questions arise when considering how to evaluate this hypothesis, including:

\begin{table}[t]
    \centering
    \caption{Evaluating Appraisal+Metareasoning Approach against requirements.}
    \begin{tabularx}{\linewidth}{cX}
    \hline
    Requirement & Analysis of Appraisal+Metareasoning Approach \\ \hline \hline
    R1 & Appraisals (especially control/power) allow agents to identify high-threat situations and focus response on fast, safe-seeking actions.\\ 
   R2 & Control and power inform what an agent might and might not be able to change in their environment. \\ 
  R3 & Goal-conduciveness and fine-grained novelty %
direct attention and support the agent in managing complex, interacting priorities at scale. \\ 
    R4  & Appraisals (especially novelty and goal-conduciveness) help an agent distinguish between uncertain outcomes and unknown situations.\\ 
    R5 & Context-mediated choice of decision strategies (including reflection) enables the agent to update policies and elaborations with experiences.\\ 
   R6 & Appraisals (especially novelty, urgency, and goal-conduciveness) inform what action and decision choices are apt for the specific moment.\\ 
   R7 & Appraisals guide choice of responses to OODS situations, within the context of tasks and goals/mission.\\ 
  R8 & Metareasoning can invoke post-hoc analysis of OODS responses (``reflection''), enabling offline learning for future instances of similar situations.\\ \hline

    \end{tabularx}
    \label{tab:appraisal_vs_requirements}
\end{table}

\begin{enumerate}
    \item Algorithms and implementations of individual appraisal dimensions. Should fixed \textit{architectural} processes generate appraisals or should generation be mediated by agent knowledge?  Human appraisal theory suggests a mix, while most computational implementations of appraisal assume one modality.
    \item Mapping from variable instances to decision strategies. Many candidates for decision-making strategies exist, but less documented  (in terms of appraisals) is choosing which decision strategies are appropriate for what situations. Creating such a mapping is a learning problem, enabling agents to tune metareasoning to specific appraisals. A key test of the hypothesis is whether this knowledge transfers effectively to other domains.
    \item Implementation architecture and integration. In Fig.~\ref{fig:compuational-analysis-picture} and~\ref{fig:appraisal_approach}, OODS processing is presented as separate from task-focused decision making. However, it is not yet clear whether appraisal processing should be a separate, parallel ``thread'' of cognitive processing or interleaved with other cognitive steps. Especially for appraisal dimensions mediated by knowledge, a more unified, interleaved approach may offer benefits, especially for agent learning. 
    \item Empirical methods: As others have observed \cite{khetarpal_towards_2022,goel_novelgym_2024_local}, test and evaluation of agents in OODS situations is challenging. Alongside new solutions, we anticipate the need for new methods designed to identify and to generate OODS situations across all three categories introduced in Table~\ref{tab:oods_situation_types}.

\end{enumerate}

\begin{credits}
\subsubsection{\ackname} This work was supported by the Office of Naval Research, contract N00014-22-1-2358. The views and conclusions contained in this document are those of the authors and should not be interpreted as representing the official policies, either expressed or implied, of the Department of Defense or Office of Naval Research. The U.S. Government is authorized to reproduce and distribute reprints for Government purposes notwithstanding any copyright notation hereon.  We would like to thank the anonymous reviewers for constructive suggestions and feedback.

\subsubsection{\discintname}
The authors have no competing interests to declare that are
relevant to the content of this article. 
\end{credits}

\bibliographystyle{splncs04}
%\bibliography{abc-zotero,format}

\begin{thebibliography}{10}
\providecommand{\url}[1]{\texttt{#1}}
\providecommand{\urlprefix}{URL }
\providecommand{\doi}[1]{https://doi.org/#1}

\bibitem{anand_prediction_2023}
Anand, N., Precup, D.: Prediction and control in continual reinforcement learning. In: Proc. of 37th {International} {Conf}. on {Neural} {Information} {Processing} {Systems}. {NIPS} '23, vol.~36, pp. 63779--63817. Curran Associates Inc. (2023)

\bibitem{anand_prediction_2023_local}
Anand, N., Precup, D.: Prediction and control in continual reinforcement learning. In: Proc. of {NeurIPS23}. vol.~36, pp. 63779--63817. Curran Associates Inc. (2023)

\bibitem{bruner_study_2017}
Bruner, J., Goodnow, J.J., Austin, G.A.: A {Study} of {Thinking}. ({Original} work published in 1956), Routledge, New York, 2 edn. (2017)

\bibitem{chen_adapt_2023}
Chen, A., Chada, G., Smith, L., Sharma, A., Fu, Z., Levine, S., Finn, C.: Adapt {On}-the-{Go}: {Behavior} {Modulation} for {Single}-{Life} {Robot} {Deployment}. In: 6th {Robot} {Learning} {Workshop}: {Pretraining}, {Fine}-{Tuning}, and {Generalization} with {Large} {Scale} {Models} (2023)

\bibitem{chen_you_2022}
Chen, A., Sharma, A., {Sergey Levine}, Finn, C.: You only live once: {Single}-life reinforcement learning. In: Proceedings of the 36th {International} {Conference} on {Neural} {Information} {Processing} {Systems}. {NIPS} '22, vol.~35, pp. 14784--14797. Curran Associates Inc., New Orleans, LA, USA (2022)

\bibitem{chen_you_2022_local}
Chen, A., Sharma, A., {Sergey Levine}, Finn, C.: You only live once: {Single}-life reinforcement learning. In: Proc. {NeurIPS22}. pp. 14784--14797. New Orleans (2022)

\bibitem{dietterich_hierarchical_2000}
Dietterich, T.G.: Hierarchical {Reinforcement} {Learning} with the {MAXQ} {Value} {Function} {Decomposition}. JAIR  \textbf{13},  227--303 (Nov 2000)

\bibitem{dumonteil_reactive_2015}
Dumonteil, G., Manfredi, G., Devy, M., Confetti, A., Sidobre, D.: Reactive {Planning} on a {Collaborative} {Robot} for {Industrial} {Applications}:. In: Proceedings of the 12th {International} {Conference} on {Informatics} in {Control}, {Automation} and {Robotics}. pp. 450--457. SCITEPRESS, Colmar, Alsace, France (2015)

\bibitem{ellsworth_appraisal_2003}
Ellsworth, P.C., Scherer, K.R.: Appraisal processes in emotion. In: Handbook of affective sciences, pp. 572--595. Series in affective science, OUP, New York (2003)

\bibitem{endsley_toward_1995}
Endsley, M.R.: Toward a {Theory} of {Situation} {Awareness} in {Dynamic} {Systems}. Human Factors  \textbf{37}(1),  32--64 (1995)

\bibitem{erol_umcp_1994}
Erol, K., Hendler, J., Nau, D.S.: {UMCP}: a sound and complete procedure for hierarchical task-network planning. In: Proc. of {AIPS}'94. pp. 249--254. AAAI Press, Chicago, Illinois (Jun 1994)

\bibitem{fraifer_autonomous_2025}
Fraifer, M.A., Coleman, J., Maguire, J., Trslić, P., Dooly, G., Toal, D.: Autonomous {Forklifts}: {State} of the {Art}—{Exploring} {Perception}, {Scanning} {Technologies} and {Functional} {Systems}—{A} {Comprehensive} {Review}. Electronics  \textbf{14}(1), ~153 (Jan 2025)

\bibitem{frijda_emotion_2009}
Frijda, N.H., Scherer, K.R.: Emotion definitions (psychological perspectives). In: Sande, D., Scherer, K.R. (eds.) The {Oxford} companion to emotion and the affective sciences, pp. 142--144. New YorkOxford University Press, New York (2009)

\bibitem{frijda_laws_2006}
Frijda, N.H.: The {Laws} of {Emotion}. Psychology Press, Mahwah, NJ, 1st edition edn. (Aug 2006)

\bibitem{gigerenzer_fast_2004_local}
Gigerenzer, G.: Fast and frugal heuristics. In: Blackwell handbook of judgment and decision making, pp. 62--88. Blackwell, Malden (2004)

\bibitem{goel_neurosymbolic_2024}
Goel, S., Lymperopoulos, P., Thielstrom, R., Krause, E., Feeney, P., Lorang, P., Schneider, S., Wei, Y., Kildebeck, E., Goss, S., Hughes, M.C., Liu, L., Sinapov, J., Scheutz, M.: A neurosymbolic cognitive architecture framework for handling novelties in open worlds. Artificial Intelligence  \textbf{331},  104111 (Jun 2024)

\bibitem{goel_novelgym_2024_local}
Goel, S., Wei, Y., Lymperopoulos, P., Churá, K., Scheutz, M., Sinapov, J.: {NovelGym}: {A} {Flexible} {Ecosystem} for {Hybrid} {Planning} and {Learning} {Agents} {Designed} for {Open} {Worlds}. In: Proc. {AAMAS}24. pp. 688--696. {AAMAS} '24 (2024)

\bibitem{gratch_domain-independent_2004}
Gratch, J., Marsella, S.: A domain-independent framework for modeling emotion. Cognitive Systems Research  \textbf{5}(4),  269--306 (Dec 2004)

\bibitem{haider_can_2024}
Haider, T., Roscher, K., Herd, B., Schmoeller~Roza, F., Burton, S.: Can you trust your {Agent}? {The} {Effect} of {Out}-of-{Distribution} {Detection} on the {Safety} of {Reinforcement} {Learning} {Systems}. In: Proceedings of the 39th {ACM}/{SIGAPP} {Symposium} on {Applied} {Computing}. pp. 1569--1578. ACM, Avila Spain (Apr 2024)

\bibitem{hansen_self-supervised_2021}
Hansen, N., Jangir, R., Sun, Y., Alenyà, G., Abbeel, P., Efros, A., Pinto, L., Wang, X.: Self-{Supervised} {Policy} {Adaptation} during {Deployment}. In: {ICLR} 2021 (2021)

\bibitem{kejriwal_challenges_2024}
Kejriwal, M., Kildebeck, E., Steininger, R., Shrivastava, A.: Challenges, evaluation and opportunities for open-world learning. Nat Mach Intell  \textbf{6}(6),  580--588 (2024)

\bibitem{khetarpal_towards_2022}
Khetarpal, K., Riemer, M., Rish, I., Precup, D.: Towards {Continual} {Reinforcement} {Learning}: {A} {Review} and {Perspectives}. JAIR  \textbf{75},  1401--1476 (Dec 2022)

\bibitem{klein_sources_1998}
Klein, G.: Sources of {Power}. The MIT Press (1998)

\bibitem{klein_making_2006}
Klein, G., Moon, B., Hoffman, R.R.: Making {Sense} of {Sensemaking} 2: {A} {Macrocognitive} {Model}. IEEE Intelligent Systems  \textbf{21}(5),  88--92 (2006)

\bibitem{laird_interactive_2017}
Laird, J.E., Gluck, K., Anderson, J.R., Forbus, K., Jenkins, O., Lebiere, C., Salvucci, D., Scheutz, M., Thomaz, A., Trafton, G., Wray, R.E., Mohan, S., Kirk, J.R.: Interactive {Task} {Learning}. IEEE Intelligent Systems  \textbf{32}(4),  6--21 (2017)

\bibitem{marinier_computational_2009}
Marinier, R.P., Laird, J.E., Lewis, R.L.: A computational unification of cognitive behavior and emotion. Cognitive Systems Research  \textbf{10}(1),  48--69 (Mar 2009)

\bibitem{marsella_computationally_2014}
Marsella, S., Gratch, J.: Computationally modeling human emotion. CACM  \textbf{57}(12),  56--67 (Nov 2014)

\bibitem{maule_effects_2000}
Maule, A.J., Hockey, G.R.J., Bdzola, L.: Effects of time-pressure on decision-making under uncertainty: changes in affective state and information processing strategy. Acta Psychologica  \textbf{104}(3),  283--301 (Jun 2000)

\bibitem{mohan_domain-independent_2024}
Mohan, S., Piotrowski, W., Stern, R., Grover, S., Kim, S., Le, J., Sher, Y., De~Kleer, J.: A domain-independent agent architecture for adaptive operation in evolving open worlds. Artificial Intelligence  \textbf{334},  104161 (Sep 2024)

\bibitem{nau_applications_2005}
Nau, D., {Tsz-Chiu Au}, Ilghami, O., Kuter, U., Munoz-Avila, H., Murdock, J., Wu, D., Yaman, F.: Applications of {SHOP} and {SHOP2}. IEEE Intelligent Systems  \textbf{20}(2),  34--41 (Mar 2005)

\bibitem{book-newell-simon}
Newell, A., Simon, H.A.: Human problem Solving. Prentice-Hall, Englewood Cliffs, NJ (1972)

\bibitem{ntsb_collision_2019}
NTSB: Collision {Between} {Vehicle} {Controlled} by {Developmental} {Automated} {Driving} {System} and {Pedestrian},{Tempe}, {Arizona}, {March} 18, 2018. Tech. Rep. Highway Accident Report NTSB/HAR-19/03, NTSB, Washington, DC. (Nov 2019)

\bibitem{roseman_cognitive_1984}
Roseman, I.J.: Cognitive determinants of emotion: {A} structural theory. In: Shaver, P. (ed.) Review of personality and social psychology, vol.~5, pp. 11--36. Sage, Beverly Hills, CA (1984)

\bibitem{russell_artificial_1995}
Russell, S., Norvig, P.: Artificial {Intelligence}: {A} {Modern} {Approach}. Prentice-Hall, Upper Saddle River, NJ (1995)

\bibitem{scherer_studying_1993}
Scherer, K.R.: Studying the emotion-antecedent appraisal process: {An} expert system approach. Cognition and Emotion  \textbf{7}(3-4),  325--355 (May 1993)

\bibitem{scherer_emotions_2009}
Scherer, K.R.: Emotions are emergent processes: they require a dynamic computational architecture. Philosophical Transactions of the Royal Society B: Biological Sciences  \textbf{364}(1535), ~3459 (Dec 2009)

\bibitem{scherer_emotion_2019}
Scherer, K.R., Moors, A.: The {Emotion} {Process}: {Event} {Appraisal} and {Component} {Differentiation}. Annual Review of Psychology  \textbf{70},  719--745 (Jan 2019)

\bibitem{shortland_choice_2020}
Shortland, N., Alison, L., Thompson, L., Barrett-Pink, C., Swan, L.: Choice and consequence: {A} naturalistic analysis of least-worst decision-making in critical incidents. Memory \& Cognition  \textbf{48}(8),  1334--1345 (Nov 2020)

\bibitem{sutton_reinforcement_2018}
Sutton, R., Barto, A.G.: Reinforcement {Learning}: {An} {Introduction}. MIT Press, Cambridge, MA, second edn. (2018)

\bibitem{wong_resilient_2018}
Wong, K.W., Ehlers, R., Kress-Gazit, H.: Resilient, {Provably}-{Correct}, and {High}-{Level} {Robot} {Behaviors}. IEEE Transactions on Robotics  \textbf{34}(4),  936--952 (Aug 2018)

\bibitem{yang_designing_2020_local}
Yang, L., Sun, Q., Ye, Z.S.: Designing {Mission} {Abort} {Strategies} {Based} on {Early}-{Warning} {Information}. IEEE Trans Indus Inform  \textbf{16}(1),  277--287 (2020)

\end{thebibliography}

%

\end{document}